\documentclass[10pt, conference, compsocconf]{IEEEtran}

\pdfoutput=1

\IEEEoverridecommandlockouts 

\usepackage{cite}

\usepackage{hyperref}

\ifCLASSINFOpdf
\else
\fi

\usepackage{epsfig}
\usepackage{graphicx}
\usepackage{amsmath}

\usepackage{subcaption}
\usepackage{bm}

\usepackage{calrsfs}
\DeclareMathAlphabet{\pazocal}{OMS}{zplm}{m}{n}

\begin{document}

%
\title{A Pyramid CNN for Dense-Leaves Segmentation}


\author{\IEEEauthorblockN{Daniel D. Morris}
\IEEEauthorblockA{
Dept. of Electrical and Computer Engineering,\\
Michigan State University,\\
East Lansing, MI 48824, USA \\
Email: dmorris@msu.edu}
}

\IEEEpubid{\makebox[\columnwidth]{} \hspace{\columnsep}\makebox[\columnwidth]{\hfill Copyright~2018 IEEE}} 

\maketitle


\begin{abstract}

Automatic detection and segmentation of overlapping leaves in dense foliage can be a difficult task, particularly for leaves with strong textures and high occlusions.  We present Dense-Leaves, an image dataset with ground truth segmentation labels that can be used to train and quantify algorithms for leaf segmentation in the wild.  We also propose a pyramid convolutional neural network with multi-scale predictions that detects and discriminates leaf boundaries from interior textures.  Using these detected boundaries, closed-contour boundaries around individual leaves are estimated with a watershed-based algorithm.  The result is an instance segmenter for dense leaves.  Promising segmentation results for leaves in dense foliage are obtained.

\end{abstract}




\section{Introduction}
\label{sec:intro}

Finding and identifying plant leaves in imagery is an area of growing research interest with significant agricultural applications.  These applications include automated inspection of crops and trees for growth assessment, pathogen detection, weed detection and species identification.  Currently these tasks often require specialized scanning devices to determine leaf shape, or significant manual effort such as placing leaves on a white background before imaging~\cite{kumar2012leafsnap}.

Progress has been made in automating leaf segmentation in controlled settings where knowledge of context aids in the task.  In~\cite{kumar2012leafsnap}, a single leaf is placed on a plain background, and can be assumed to flat and fully visible.  In~\cite{GrandBrochier:2015:treeLeaves} a single tree leaf is known to be in the center of each image. In~\cite{Minervini201435,MINERVINI:2016}, leaves of known type are grown in a lab with the plant stem near the center of the image.  However, for robotic outdoor inspection applications, leaves are likely to be against varying backgrounds and any number may occur anywhere in an image.  For outdoor applications single leaf-in-image segmentation methods will not suffice.  That motivates the dataset in this paper, which focuses on leaves in the wild; namely images containing a large number of overlapping leaves in arbitrary locations and varying backgrounds.  Segmenting leaves in this type of environment is a necessary step towards automated plant inspection in unstructured environments.

\begin{figure}
\begin{center}
\begin{minipage}[b]{0.5\linewidth}
\includegraphics[bb=0 0 1000 1000,trim=180 295 180 290,clip,width=0.98\linewidth]{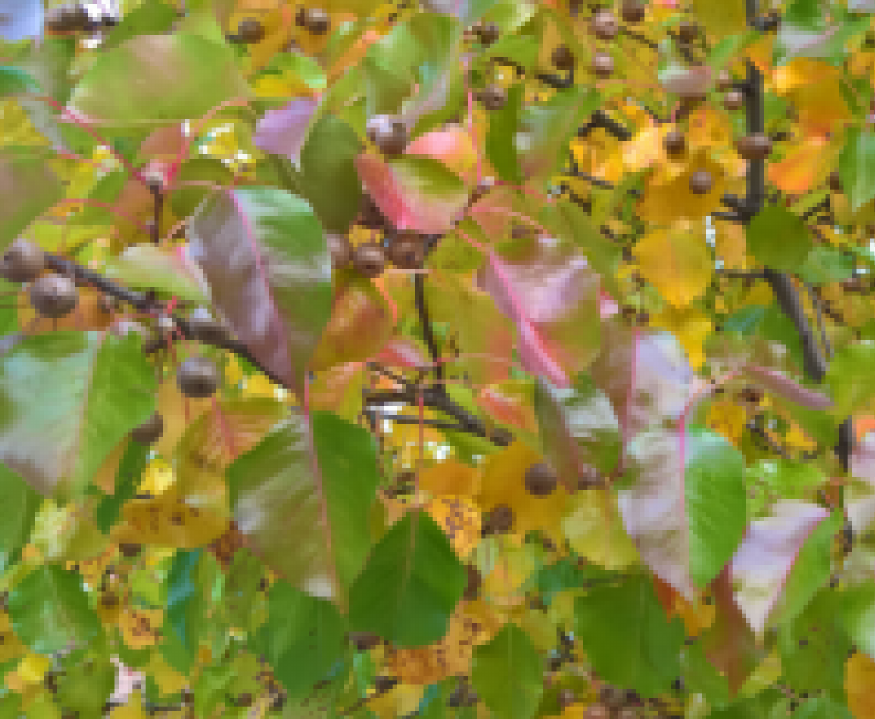}
\subcaption{}\label{fig:ex1a}
\end{minipage}%
\begin{minipage}[b]{0.5\linewidth}
\includegraphics[bb=0 0 1000 1000,trim=180 295 180 290,clip,width=0.98\linewidth]{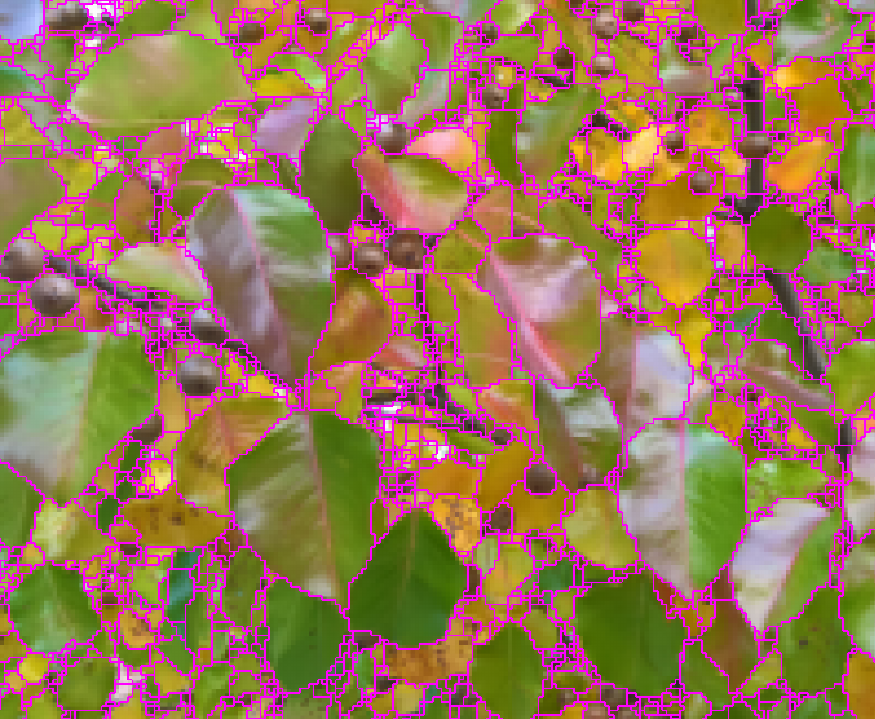}
\subcaption{}\label{fig:ex1b}
\end{minipage}
\caption{(a) An example dense foliage image in which we seek to segment individual leaves. (b) Closed boundary segments output by our proposed method where the visible portion of each leaf is ideally bounded by a single segment. }
\label{fig:Example1}
\end{center}
\end{figure}

The leaf segmentation goal differs somewhat from other segmentation tasks.  On one hand leaves have a large variation in shape, size and appearance with sometimes complex boundaries.  This makes the task more than just an object detection, as the existence of a leaf gives little information on its boundary.  On the other hand identifying all pixels belonging to the target class, as is the case for some segmentation problems, does not solve the problem as this will merge overlapping leaves.  And finally, boundary detection on its own is insufficient as gaps will result in leaves being merged.  In this work, images are acquired from a single viewpoint, and occluded regions are unknown. Thus we consider leaf segmentation as finding closed boundaries around contiguous, visible portions of leaves.  

\IEEEpubidadjcol

Automated boundary detection has been an area of active research.  Statistical feature matching methods~\cite{Konishi:2003,arbelaez:2011,Dollar:2015} discriminate meaningful edge boundaries from apparent boundaries.   The segmentation method by Arbelaez et al.~\cite{arbelaez:2011} uses learned discriminators to find semantically meaninful image boundaries.  This has advanced futher in the era of convolutional neural networks to by additional boundary methods~\cite{Shen:2015,Xie:HED:2017} learned over large datasets.  Notably~\cite{Xie:HED:2017} leverages object recognition from a VGG-16 deep network~\cite{Simonyan:2014} to guide boundary multi-resolution boundary detection.  

These methods have been driven by the BSD~\cite{Martin:BSD:2001} and NYU~\cite{Silberman:ECCV:2012} datasets of natural objects.  Results in this paper show that our task of segmenting overlapping leaves cannot rely on methods trained on these datasets, but that a new dataset is needed.  Here we both present such a dataset and propose a convolutional neural network architecture that can capture subtle differences in overlapping leaves and detect their boundaries.  In addition a boundary completion method similar to that in~\cite{arbelaez:2011}, but with scale independence, is proposed to obtain individual leaf segmentation.

  \section{Related Work}
\label{sec:related}

There has been significant work in color-image based leaf segmentation.  Approaches include active polygons~\cite{Cerutti2011,Rabatel2001} and active contours~\cite{mishra2011decoupled,Qiangqiang2015}.  A number of signatures from simple background models to shape priors and color signatures~\cite{Wang2008, Bai2011} have been used for detection.  In~\cite{Pape2015} color histograms are used to separate leaves from the background, and partially overlapping leaves are split using boundary shape cues.  The LeafSnap project~\cite{kumar2012leafsnap,Soares2013} built an automated leaf segmentation method using a variety of cues.  To be robust required that leaves be individually placed against a plain background.  Level set methods, such as~\cite{WangX2012} can segment overlapping leaves, but require a human to input a principal line of symmetry.  More recently a dataset of tree leaves is collected~\cite{GrandBrochier:2015:treeLeaves}, and an evaluation of segmentation methods.  In these images, the leaves are centered in the image and viewed flat on without occlusion.  In our application the problem is much more complicated with varying location and overlap.

Convolutional neural networks (CNNs) have begun to be used in plant analysis.  Deep CNNs have been proposed for overhead leaf counting~\cite{Aich2017_LeafCounting,Dobrescu2017_DeepLeafCounting} of potted plants.  These leverage deep networks including ResNet50~\cite{He2016_ResNet50}.
Finding plant spikes in images~\cite{Pound2017_Spikes} uses the hourglass networks~\cite{Newell2016_StackedHourglass} to generate a heatmap predicting plant feature locations.  More complex structures such as plant centers observed from overhead imagery have been trained and located using multiple instance learning~\cite{Chen2017_LocatingCropPlantCenters}.  

This work proposes a pyramid convolutional network that seeks to bring together advances in statistical boundary detection~\cite{Konishi:2003,arbelaez:2011,Dollar:2015} and convolutional neural network boundary detection~\cite{Shen:2015,Xie:HED:2017} and new CNN architectures~\cite{Newell2016_StackedHourglass} to tackle the challenging task segmenting dense-leaves observed not in the lab but in the wild.  A new Dense-Leaves dataset is provided for training and testing for this challenging problem.

\section{Dataset}
\label{sec:data}

\begin{figure}
\begin{center}
\begin{minipage}[b]{0.45\linewidth}
\includegraphics[bb=0 0 1000 1000,trim=150 274 150 270,clip,width=0.98\linewidth]{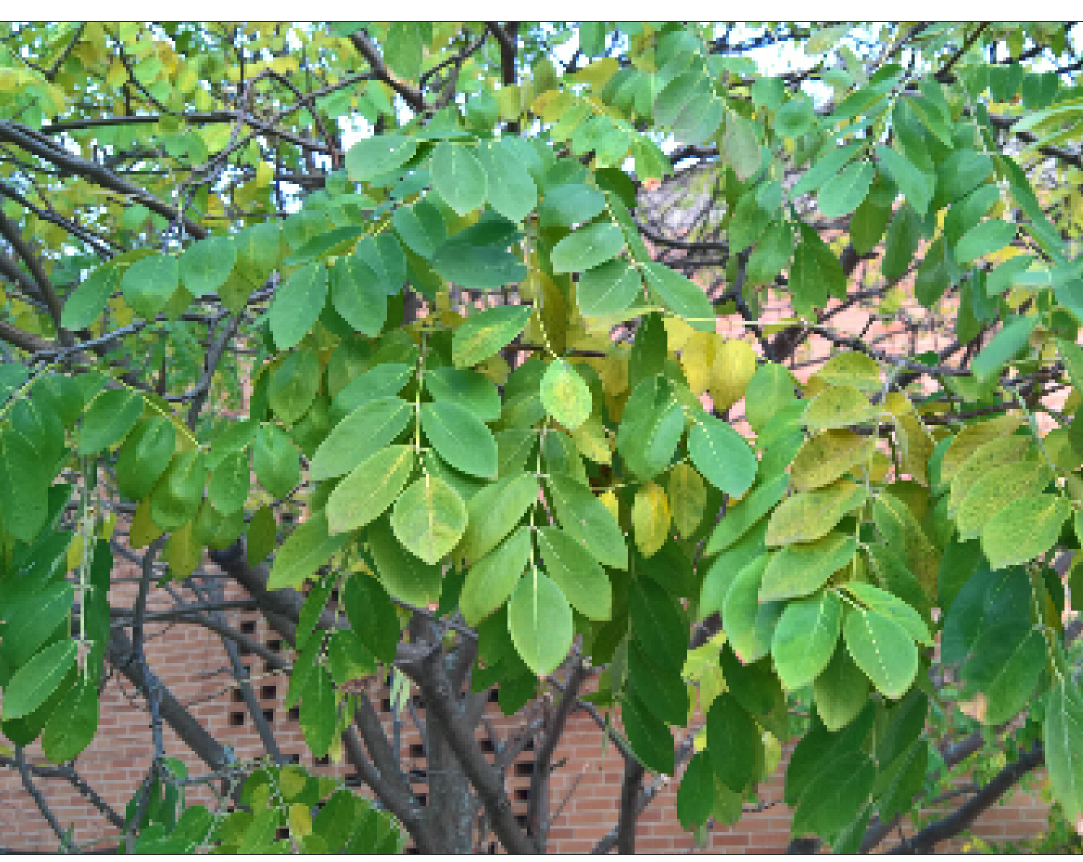}
\subcaption{}\label{fig:ima}
\end{minipage}%
\begin{minipage}[b]{0.45\linewidth}
\includegraphics[bb=0 0 1000 1000,trim=150 274 150 270,clip,width=0.98\linewidth]{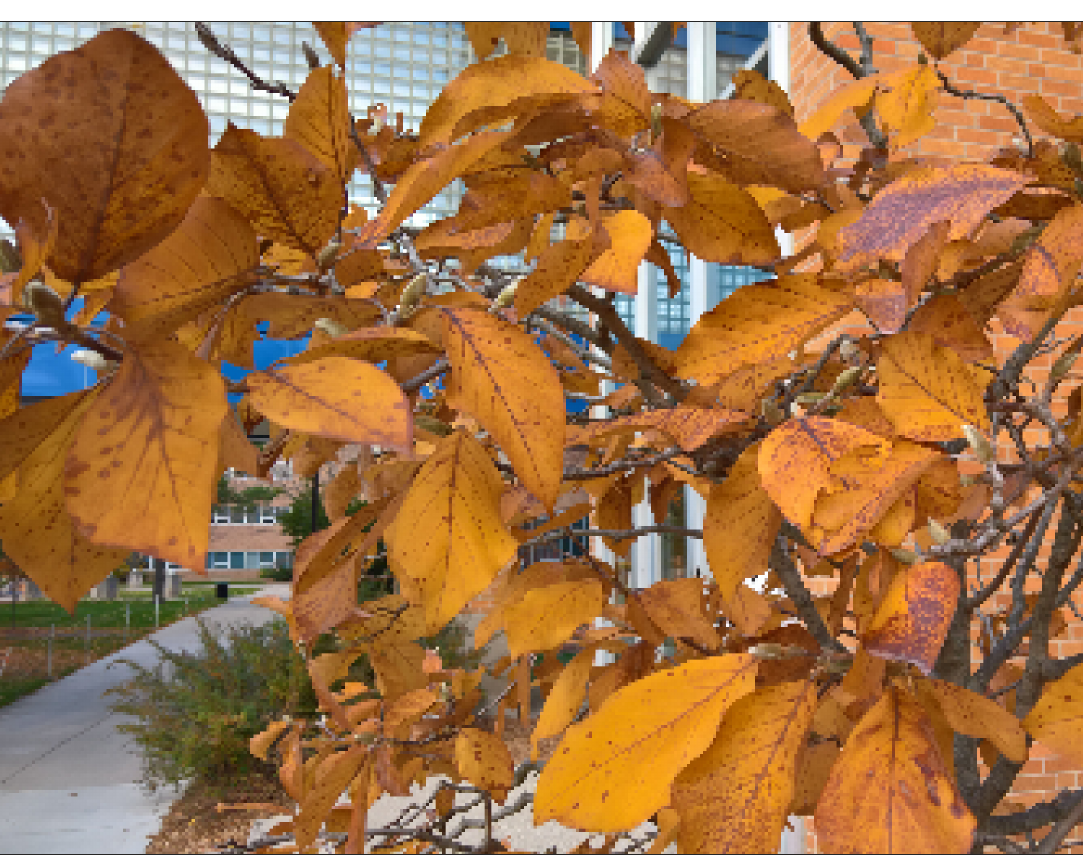}
\subcaption{}\label{fig:imb}
\end{minipage}
\begin{minipage}[b]{0.45\linewidth}
\includegraphics[bb=0 0 1000 1000,trim=150 274 150 270,clip,width=0.96\linewidth]{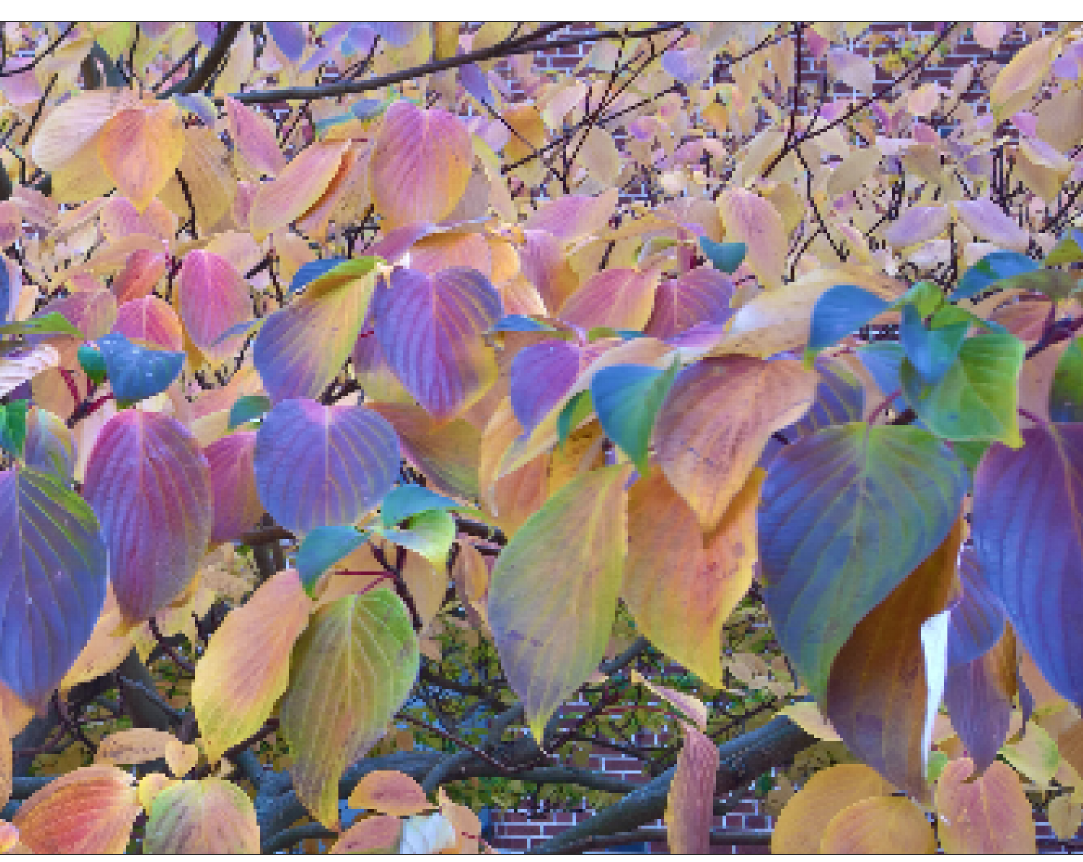}
\subcaption{}\label{fig:imc}
\end{minipage}
\begin{minipage}[b]{0.45\linewidth}
\includegraphics[bb=0 0 1000 1000,trim=150 274 150 270,clip,width=0.96\linewidth]{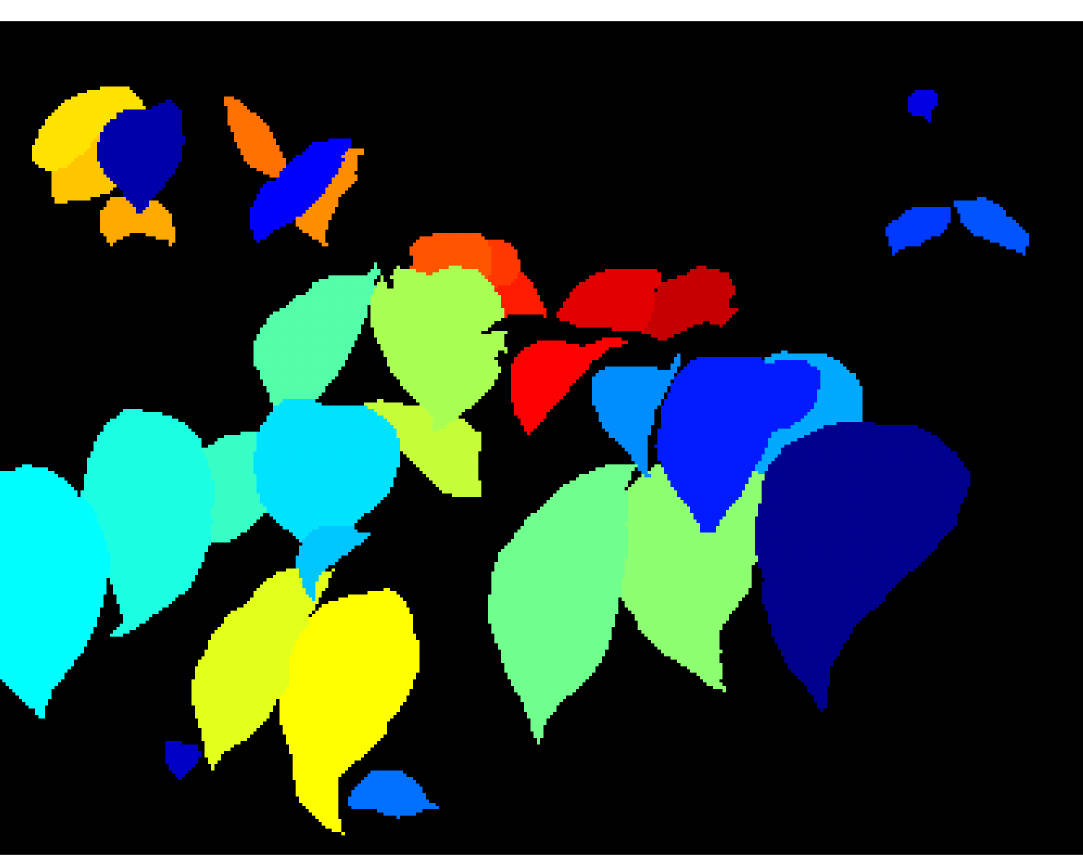}
\subcaption{}\label{fig:imd}
\end{minipage}
\caption{(a)-(c) Sample images from the dataset. (d) Human segments for (c).}
\label{fig:Images}
\end{center}
\end{figure}

To make progress in the task of instance segmentation of leaves amongst dense foliage, it is useful to have a labeled dataset appropriate to this task.  This paper presents the MSU Dense-Leaves dataset\footnote{MSU Dense-Leaves available at \url{https://www.egr.msu.edu/denseleaves/}}.  It consists of 108 images at resolution $1024\times 768$, each of dense foliage from trees, vines and bushes on or near the Michigan State University campus.   All images are under overcast skies or at night.  They are divided into a training, validation and test sets containing 73, 13 and 22 images respectively.  Some examples are shown in Figs.~\ref{fig:Example1} and \ref{fig:Images}.

Dense foliage means that there is significant leaf occlusion.  It is not possible for human labelers to distinguish all the leaves due both to high occlusion and in some cases low contrast.  Thus labelers were given the task of outlining on average 20 leaves per image, and generated a total of over 2200 ground truth boundaries.  This includes both fully visible leaves and partially visible leaves, with an emphasis on overlapping leaves to support the goal of identifying and segmenting overlapping leaves.  Polygons are converted into a labeled segmentation image with one segment or label per pixel.  Where leaves overlap, polygons may overlap, and pixels are assigned to the first or ``top'' polygon.  In this way adjacent segments will have sharp divisions with no empty pixels between them.  The portions of the images that are not overlapped by the polygons are given a label 0 indicating unknown.  It includes unlabeled leaves as well as branches, ground, sky and other background objects.

\section{Task Goals}
\label{sec:task}

\begin{figure}
\begin{center}
\begin{minipage}[b]{0.5\linewidth}
\includegraphics[bb=0 0 1000 1000,trim=180 295 180 290,clip,width=0.98\linewidth]{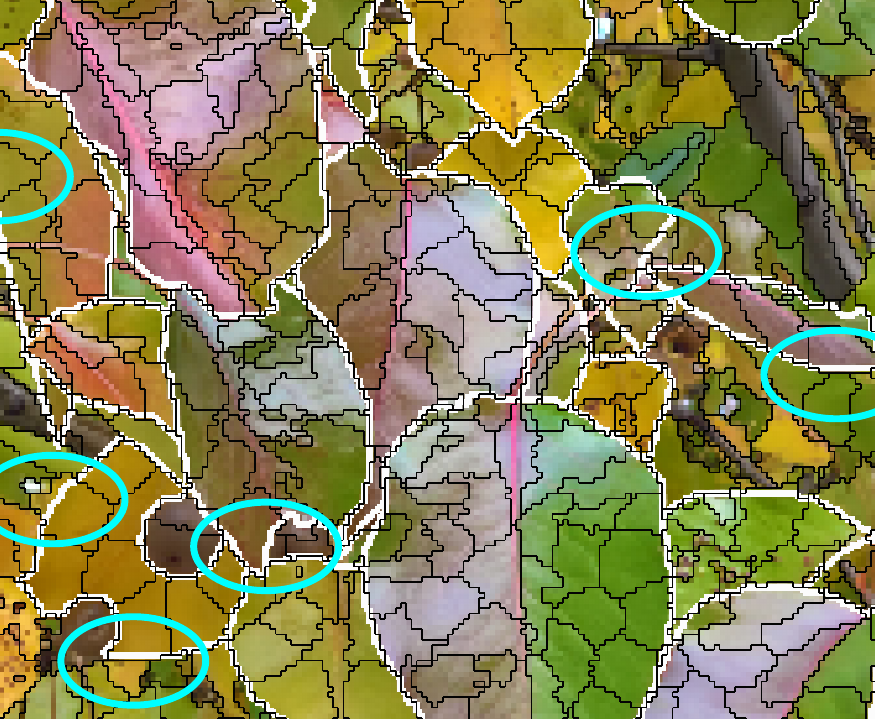}
\subcaption{}\label{fig:bnda}
\end{minipage}%
\begin{minipage}[b]{0.5\linewidth}
\includegraphics[bb=0 0 1000 1000,trim=180 295 180 290,clip,width=0.98\linewidth]{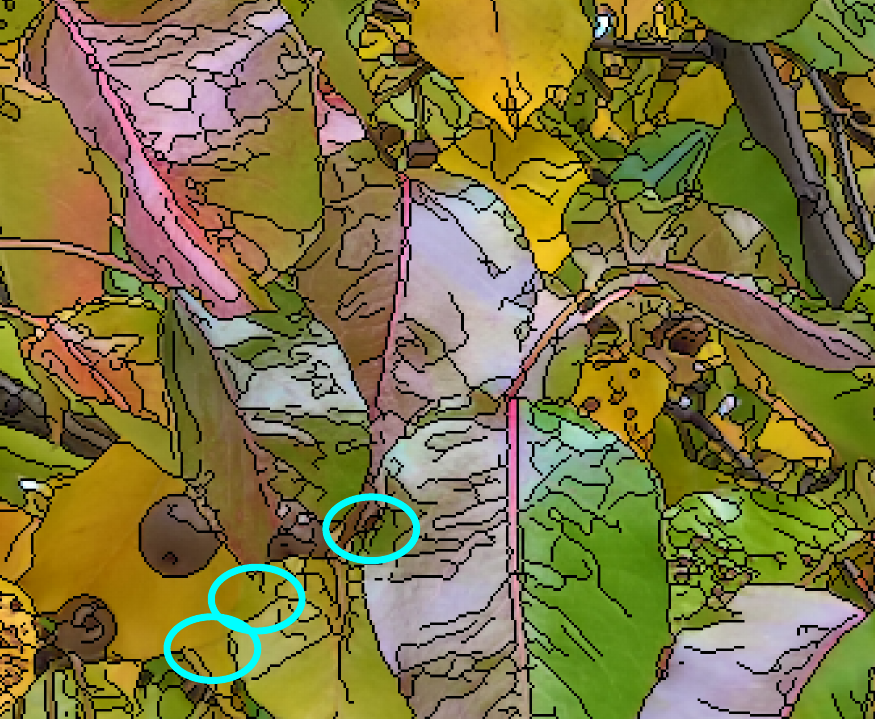}
\subcaption{}\label{fig:bndb}
\end{minipage}
\caption{A close-up of a rectangular region in Fig.~\ref{fig:Example1} showing black edges from two possible boundary detection methods: (a) superpixels~\cite{Achanta:2012} and (b) Canny edge detection~\cite{Canny:1986}.  The black lines in both cases significantly over-segment the leaves, and in some places miss the true boundaries as highlighted by the ellipses.  It is not easy to start from these boundaries and estimate the true boundaries.  This paper proposes an alternative method that avoids relying on superpixel segments in finding leaf boundaries.}
\label{fig:Example1Bounds}
\end{center}
\end{figure}

Our goal is to find closed-boundary segments that precisely outline the visible portions of the leaves in an image.  This is challenging for humans, which we use for labeling leaves, and so we limit the focus to leaves with a large, single, contiguous segment.  While we seek precise segmentations of leaves, we purposely do not address the problem of classifying leaf pixels vs. non-leaf pixels.  Leaf classification is important but its difficulty is quite problem-dependent; in some cases it can be very simple such as green leaves against a brown background.  In general case it is hard as leaf fragments could be confused with many objects and so there is a huge possible negative class.  It is also not easy to define what should be considered leaf pixels; should grass be considered leaves, or brown or decomposing leaves?  Moreover, classifying leaf from non-leaf pixels often will not help in the more difficult segmentation task where leaves are densely packed and segmentation task is to separate leaf from leaf.  Thus this paper restricts itself to finding leaf boundaries and not leaf classification.

We seek segments of the visible portion of a leaves.  While for some applications if may be preferable to obtain the outline of occluded portions, we do not address that here in part because we do not have the ground truth, but also because this is likely to be more effective when one knows the leaf type can can model the shape variation.  Here we seek to detect a wide variety of tree, bush and vine leaves.

Is leaf segmentation a difficult or an easy problem?  It depends on numerous factors including the color difference between a leaf and the background, the shape complexity including viewing angle and occlusions, internal variations in color and texture and features, and prominence of boundary edges.  There are certainly simple segmentation situations where boundary cues are prominent, but there are also numerous cases where internal edges and color variations are more significant than those on the boundary.   Fig.~\ref{fig:Example1Bounds} illustrates part of the difficulty in segmenting leaves with internal texture and color variation.  Superpixel methods are often used as starting points for object segmentation, but for complex leaves this presents two problems: (1) superpixels occasionally span the boundaries breaking the assumption of an over-segmentation, and (2) it is not clear how to merge superpixels in a way that respects leaf boundaries.  Classical edge detection methods, such as Canny~\cite{Canny:1986}, may not distinguish between leaf boundaries and internal structure.  That motivates this paper which seeks to discriminate leaf boundaries from internal structure and so enable leaf segmentation.

\section{Method}
\label{sec:method}

Leaf segmentation is partitioned into two sub-tasks: (1) finding boundary pixels and (2) obtaining contiguous segments from boundary estimates.  Our methods for each of these are described as follows.

\subsection{Boundary Detection}
\label{sec:BD} 

\begin{figure}
\begin{center}
\begin{minipage}[b]{0.31\linewidth}
\includegraphics[bb=0 0 1000 1000,trim=160 234 150 220,clip,width=0.98\linewidth]{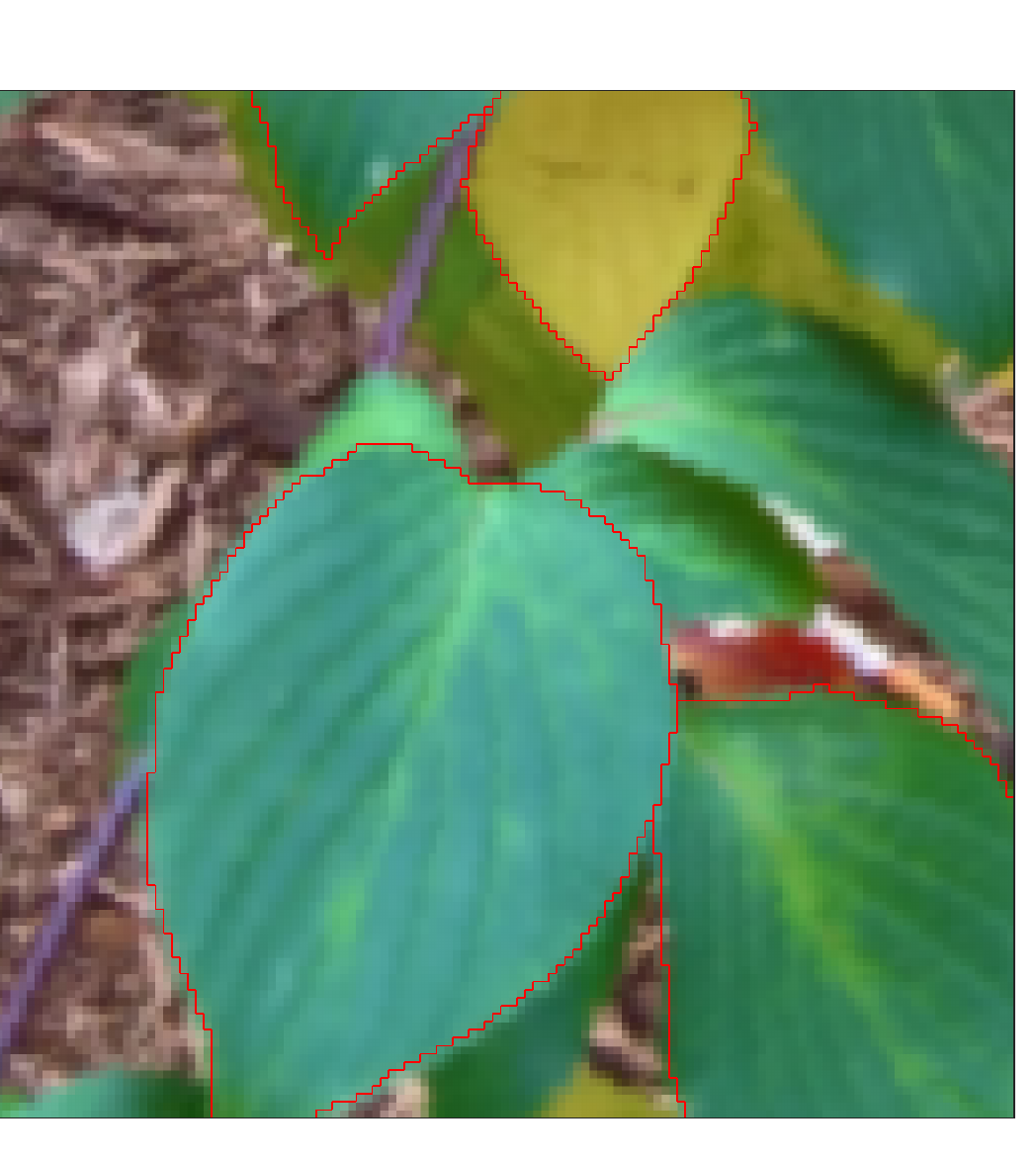}
\subcaption{}\label{fig:patcha}
\end{minipage}%
\begin{minipage}[b]{0.31\linewidth}
\includegraphics[bb=0 0 1000 1000,trim=160 234 150 220,clip,width=0.98\linewidth]{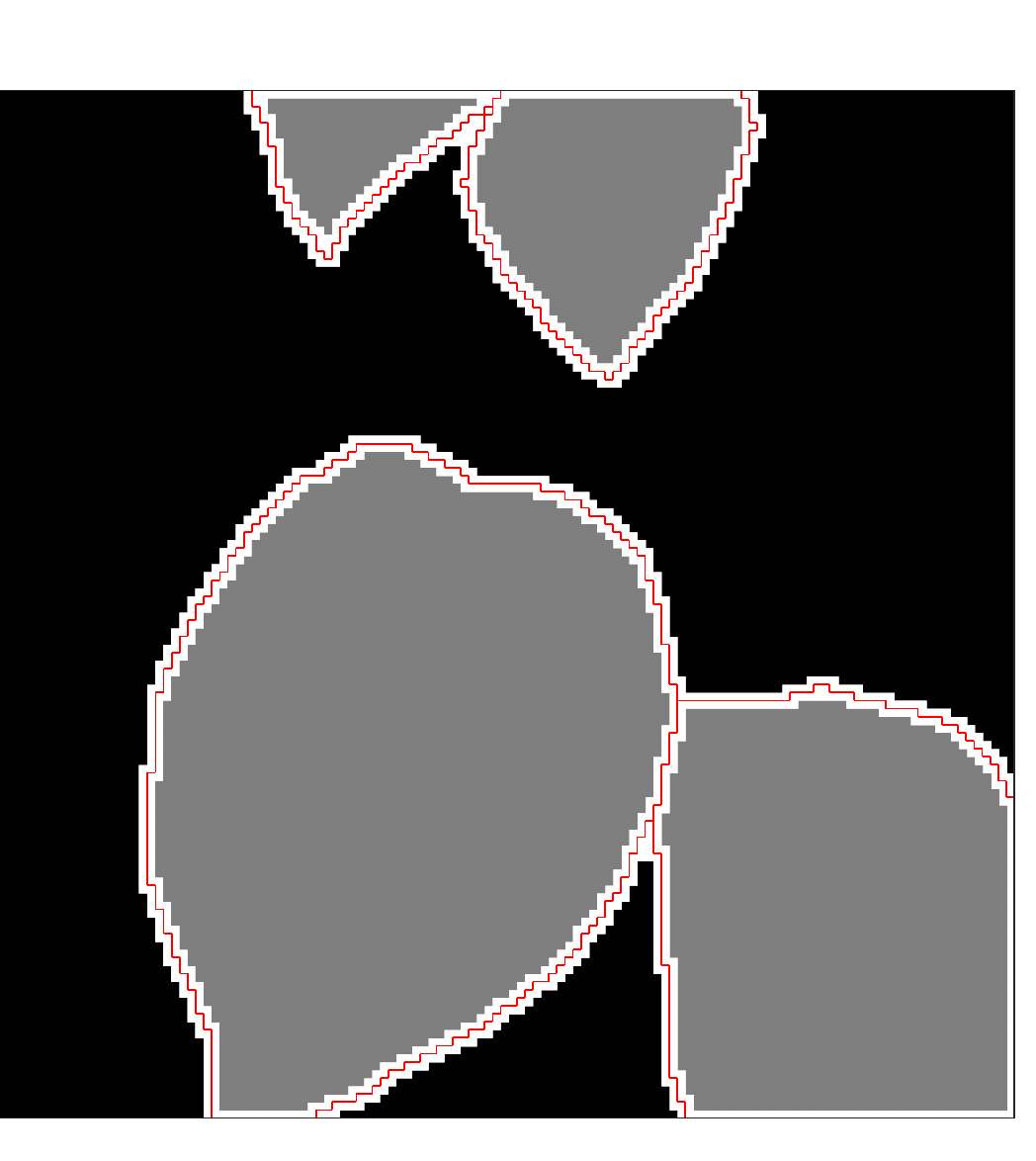}
\subcaption{}\label{fig:patchb}
\end{minipage}
\begin{minipage}[b]{0.31\linewidth}
\includegraphics[bb=0 0 1000 1000,trim=160 234 150 220,clip,width=0.98\linewidth]{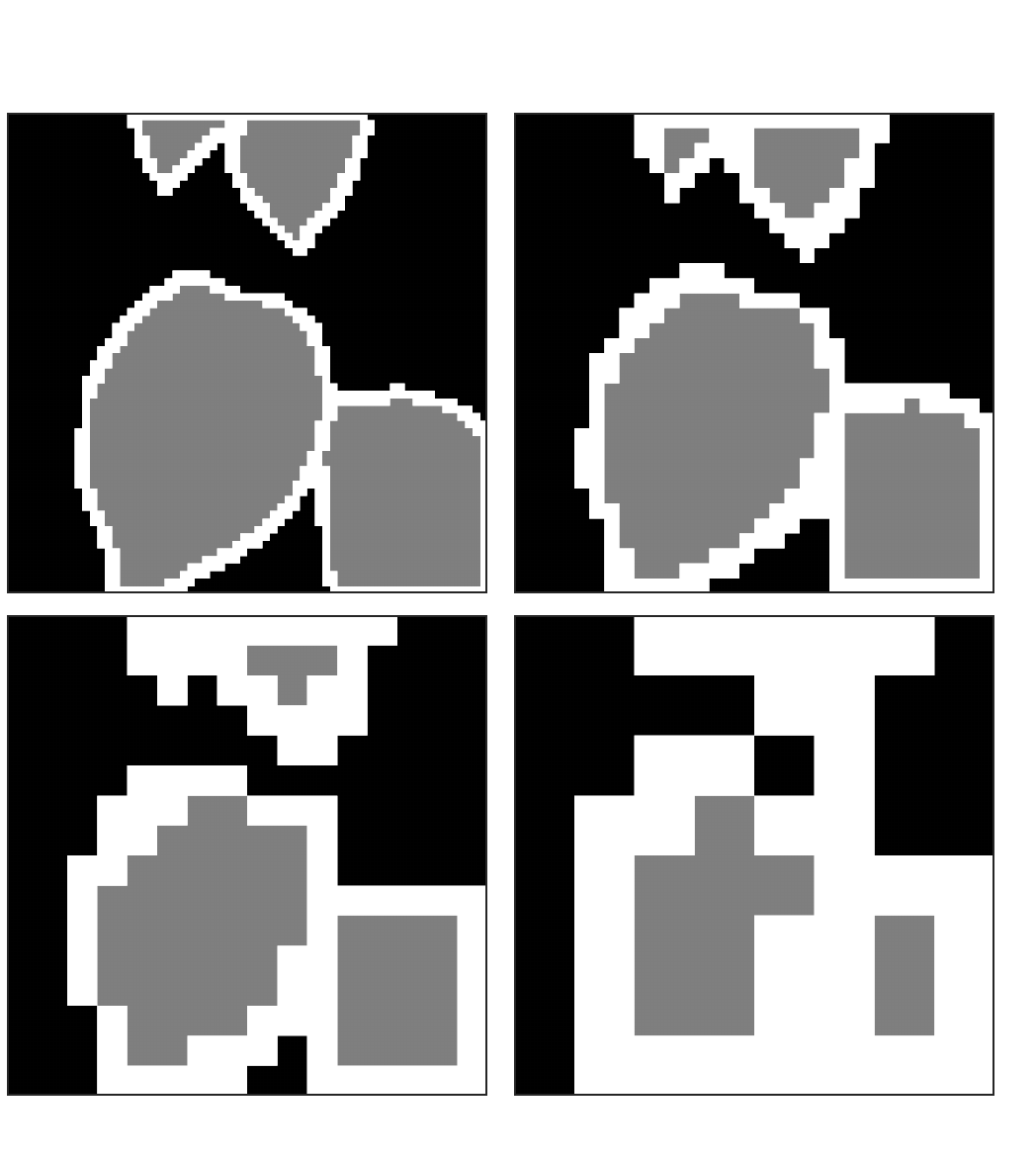}
\subcaption{}\label{fig:patchc}
\end{minipage}
\caption{(a) A training patch containing portions of 4 labeled leaves with boundaries in red.  This is converted into a binary edge (white) vs. interior (gray) classification in (b), with boundaries being 2 pixels wide.   Unlabeled regions (black) are given zero weight and do not contribute to the optimization.  Our network is trained on multi-resolution classification, and pixel classifications for the 4 lower levels are shown in (c).   }
\label{fig:Patch}
\end{center}
\end{figure}

We pose boundary detection as a binary classification problem.  This enables the use of a fully convolutional network with a simple loss function.   Given a set of leaf segments as in Fig.~\ref{fig:Patch}(a), each pixel is either a boundary or an interior pixel, as in Fig.~\ref{fig:Patch}(b).  For a given segment, its boundary pixels are those within a 1.5 pixel radius of pixels outside it.  This applies to unlabeled regions, although the interior pixels of unlabeled regions are unknown and given zero weight during learning.  This leads two a 2-pixel wide boundary around each segment. 

Our network will predict boundaries at multiple resolutions, and to train this we define boundaries at lower resolutions as follows.  Each resolution is half its parent and a pixel covers four pixels in its parent.  This pixel is declared an edge if any of the parent pixels are edges, otherwise if any of its parents are unknown it is unknown, and if neither of those conditions applies, then it is an interior pixel.  This labeling is illustrated in Fig.~\ref{fig:Patch}(c).

\subsubsection{Network Architecture}

To detect leaf boundaries it is important to gather information at both high and low resolutions.  When the color  or intensity change across a leaf boundary is small, the boundary can often be determined using contextual cues.  These cues are sometimes high-resolution texture changes and other times low-resolution extended features.     For difficult leaf segmentation tasks, human labelers are observed to examine both close-up views and low-resolution views, likely for these reasons.    Thus modeling these multi-resolution effects is likely important to robust boundary detection.

\begin{figure}
\begin{center}
\includegraphics[bb=0 0 1000 1000,trim=200 100 360 150,clip,width=\linewidth]{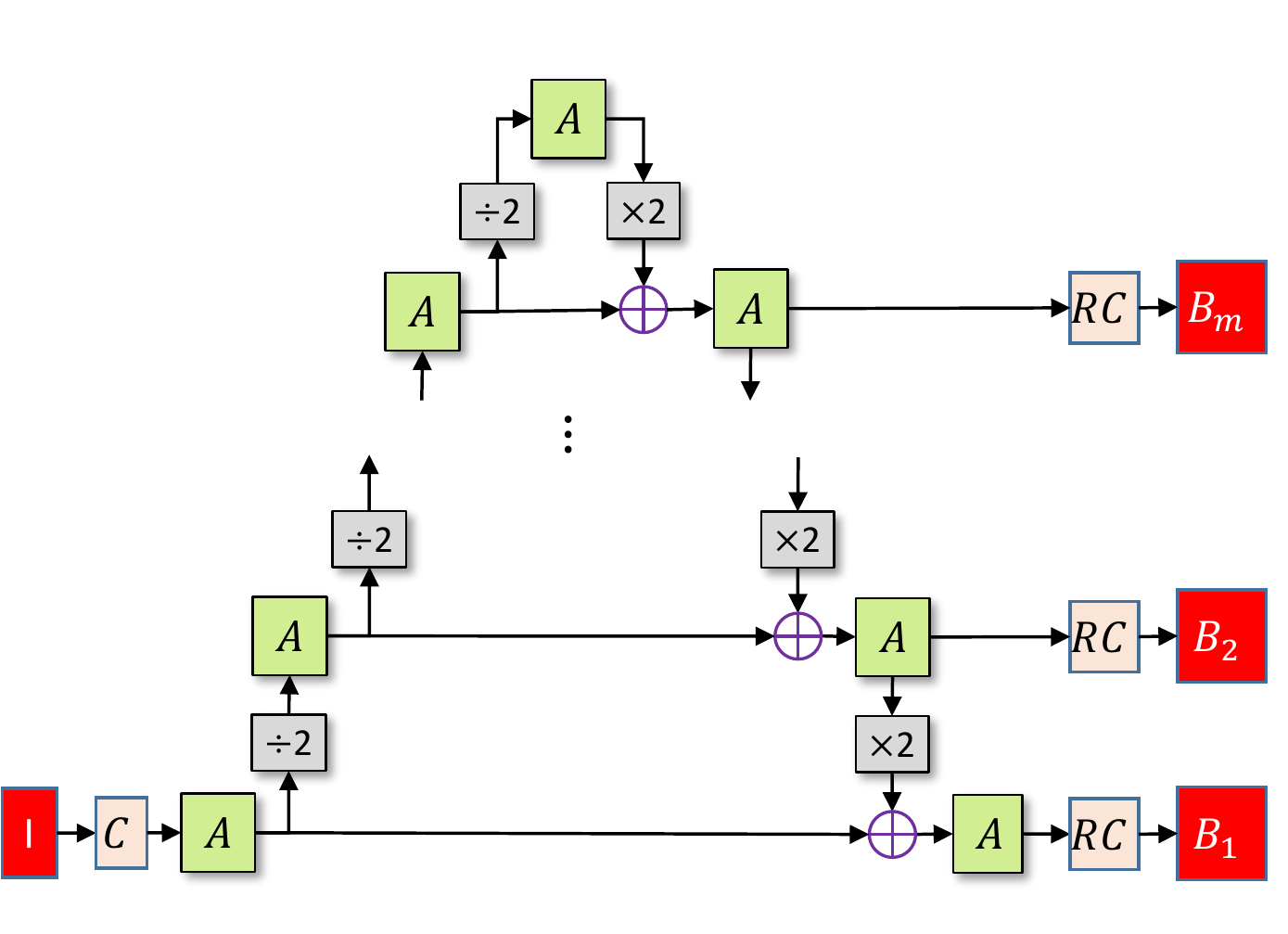} \\
($a$) \\
\includegraphics[bb=0 0 1000 1000,trim=340 280 420 190,clip,width=0.5\linewidth]{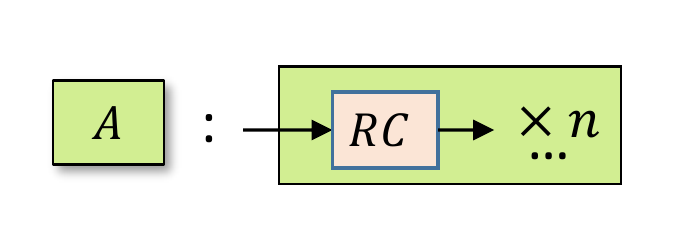}  \\
($b$)
\end{center}
\caption{(a) Our fully convolutional Pyramid Network (PN) is organized recursively in layers.  Output pixel classifications are predicted at each level, indicated by $B_i$.  Our implementation had 6 levels and 5 output predictions.   (b) A linear network component of (a).  We experimented with various number of repeats and found best performance with $n=4$.  An $RC$ indicates a ReLU followed by a $3\times3$ convolution.  The $\div2$ a two-fold down-sampling via max-pooling and the $\times2$ indicates a two-fold up-sampling using nearest-neighbor. }
\label{fig:NetArch}
\end{figure}

This motivates our fully-convolutional Pyramid Network (PN) illustrated in Fig.~\ref{fig:NetArch}.  Here a sequence of convolutions operate at successive resolutions as information flows up the pyramid to the lowest resolution.  Then again convolutions operate as the information flows down each layer to highest resolution.  This enables features at high and low resolutions to influence each other.  In addition, information flows across each level and predicts an edge classification for that resolution level.  This network structure is quite similar to the hourglass network in~\cite{Newell2016_StackedHourglass}.  The main difference is the prediction at each resolution level.  These predictions provide a number of advantages.  Losses at each prediction shorten the back propagation path leading to faster convergence.  They give explicit multi-scale structure to the network, and we can use the output predictions at multiple resolutions in the next step of our algorithm.  

A fixed number of 32 channels was used throughout the network.  The network was implemented in MatConvNet~\cite{vedaldi:2015:matconvnet}.

\subsubsection{Patch selection and data augmentation}

Due to GPU memory limitations, training is performed on batches of $128\times 128$-pixel patches, as illustrated in Fig.~\ref{fig:Patch}.  It is not necessary that leaves be fully contained in a patch, as the patch boundary effectively models an occluding edge.  Patches are selected by tiling over the training images.  Since edge detection is likely to be sensitive to edge orientation, we augmented the dataset by rotating each image every 15 degrees between 0 and 90 deg, and are flipped horizontally and vertically.

\subsubsection{Pixel loss and weighting}

A softmax loss is used to specify a cross entropy function on edge and interior pixels.  Since the ground truth boundary pixels are significantly outnumbered by leaf interior pixels, following~\cite{Xie:HED:2017}, the boundary and interior pixels are reweighted to give them equal overall influence.  Unlabeled pixels are given zero weight.  One issue with this is that both easy and difficult interior pixels are given equal weight.  We found that we could improve performance by periodically re-weighting; giving additional weight to the 10\% lowest-scoring edge pixels and 10\% interior pixels.  This hard-negative mining was performed every 5 epochs.  

%
%

\subsection{Segment Building}
\label{sec:building}

The result of boundary detection is a density image giving a measure of the probability of edge vs. non-edge.  We seek to turn this into a set of contiguous segments.  We assume that the boundary detection result will suppress the majority of the internal edges while recovering the majority of the boundary edges.   The goal here is to find segments that connect internal pixels without leaking through gaps left by imperfect boundary detection.   Simply thresholding this and performing connected components will produce contiguous segments, but if there are any gaps in the leaf boundaries this will merge leaves with their adjacent segments resulting in a fragile and often poor segments.    

Our method has two steps.  An initial segment creation step followed by segment merging.  In segment creation we seek an over-segmentation with boundaries that include the leaf boundaries.  This is achieved by performing watershed using a Quickshift~\cite{Vedaldi:Quick:2008} on the predicted edge probability image.  Segment merging is more involved as we seek to avoid merging segments across leaf boundaries which may have gaps.  We define the affinity between 2 pixels $i$, $j$ as:
\begin{equation}
a(i,j) = 1 - max(p_i,p_j)
\label{eq:aff}
\end{equation}
where $p_i$ and $p_j$ are the probability of edge for the pixels $i$ and $j$.  This is obtained directly from the predicted edge image.  The affinity from segment $s_a$ to segment $s_b$ is the total affinity across their boundary divided by the total affinity of all boundary pixels of $s_a$.  Using this directional affinity measure, segments are connected to larger segments (which tend to be closer to the center of leaves) with greatest affinity as long as this affinity is greater than threshold $t$.  A value of $t=0.1$ gave good results and was used throughout.  Using this rule, trees are formed over the segments and all members of a tree are merged.  This is repeated until the number of segments does not change.  The result is a set of closed segments in which segments tend to merge with others within the same leaf and which will in many cases lead to a single segment per leaf. In some cases this can fail to fully merge segments within leaves when the boundaries between segments are small compared to the segment circumference as occurs in some occlusions.  Merging these may require models of leaf shapes.

\section{Results}
\label{sec:results}

To  evaluate and compare boundary detection methods, we propose an average precision score as follows.  Edge pixels are defined on the edges of segments as in Section~\ref{sec:BD}.  The score for each edge pixel is the maximum score in a 2.5 pixel radius around it, enabling compatibility with both thin and wide edge detection methods. This is important as the thickness of a boundary edge is not so important as its coverage.  Those pixels within a labeled segment and not within this radius of an edge pixel are considered interior pixels.  A precision-recall curve can be calculated for any boundary detection method that gives a score to each pixel, and the average precision (Boundary AP) calculated.   Results for this measure are shown table~\ref{table:results}.

To evaluate the final image segmentations, we must associate estimated segments with ground truth segments.  Each ground truth segment should have a single, unique estimated segment associated with it.  Then how well these segments cover and/or extend beyond the ground truth segments can be measured.  We perform this association by maximizing the dice score defined for a given segment as:
\begin{equation}
dice = \frac{2PR}{P+R}
\label{eq:dice}
\end{equation}
where $P$ is the precision and $R$ the recall for pixels belonging to the segment.  The association that maximizes the sum of dice scores for ground truth segments is selected.  Then given this association the total precision, recall and dice over all pixels is calculated, and these are reported in table~\ref{table:results}.

The PN network was trained on the Dense-Leaves training dataset and evaluated on the test images.  Three different variants were evaluated.  In the first, PN-P1, only a single prediction output at the highest resolution was used to create an optimization loss.  This had poorer performance than the second network, PN-P5-S5, which trained with losses on each of the 5 predictions.  Here the outputs are fused into a single image by upsampling the lowest resolution and summing with the next higher resolution, and repeating this until all images have been added to the highest resolution image, and then divided by the number of resolutions. The final variant, PN-P5-S1, trained with five predictions but only used the highest resolution prediction for segmentation.  All of the variants used the highest-resolution prediction to estimate the boundary pixels.  Results and comparisons are in Table~\ref{table:results}.

\begin{figure}
\begin{center}
\begin{minipage}[b]{0.5\linewidth}
\includegraphics[bb=0 0 1000 1000,trim=180 295 180 290,clip,width=0.98\linewidth]{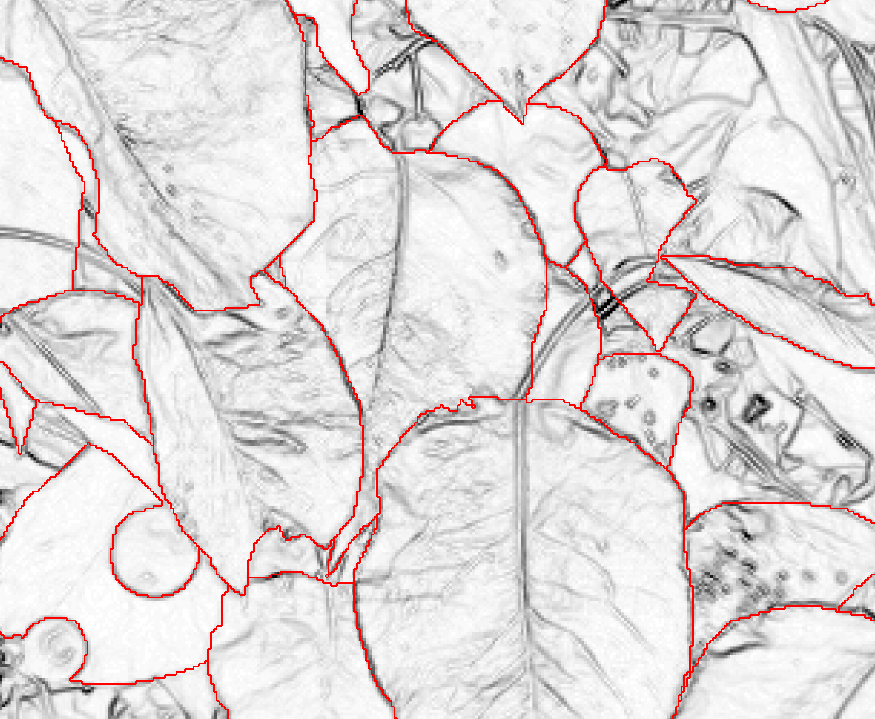}
\subcaption{}
\end{minipage}%
\begin{minipage}[b]{0.5\linewidth}
\includegraphics[bb=0 0 1000 1000,trim=180 295 180 290,clip,width=0.98\linewidth]{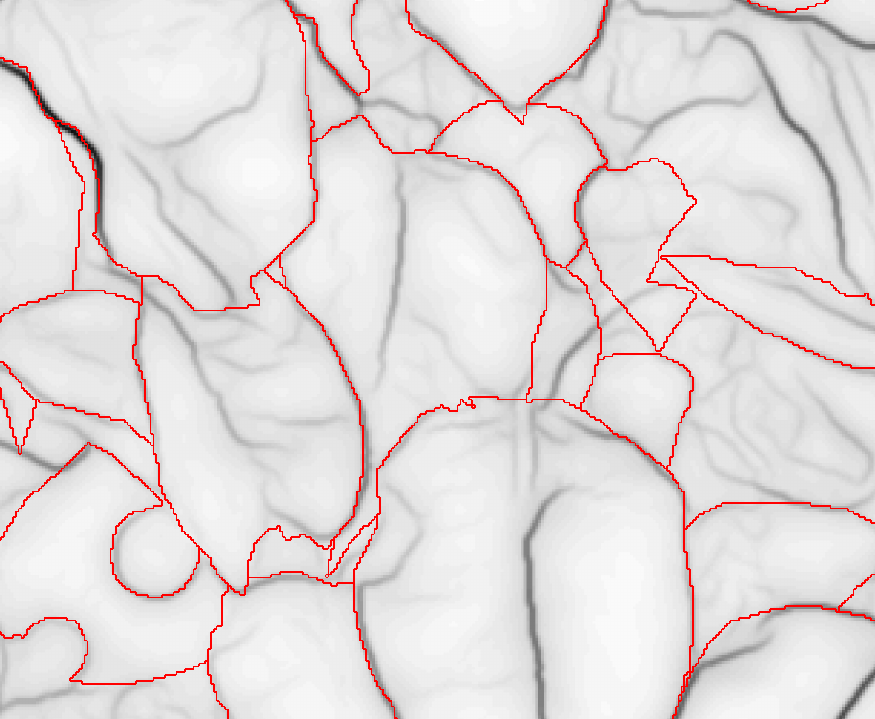}
\subcaption{}
\end{minipage}
\begin{minipage}[b]{0.5\linewidth}
\includegraphics[bb=0 0 1000 1000,trim=180 295 180 290,clip,width=0.98\linewidth]{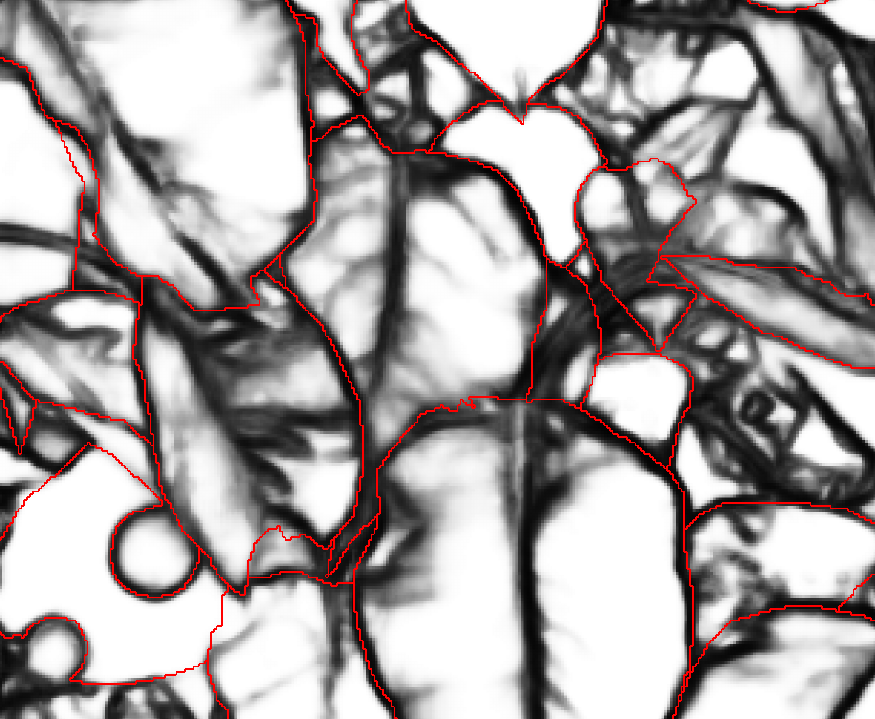}
\subcaption{}
\end{minipage}%
\begin{minipage}[b]{0.5\linewidth}
\includegraphics[bb=0 0 1000 1000,trim=180 295 180 290,clip,width=0.98\linewidth]{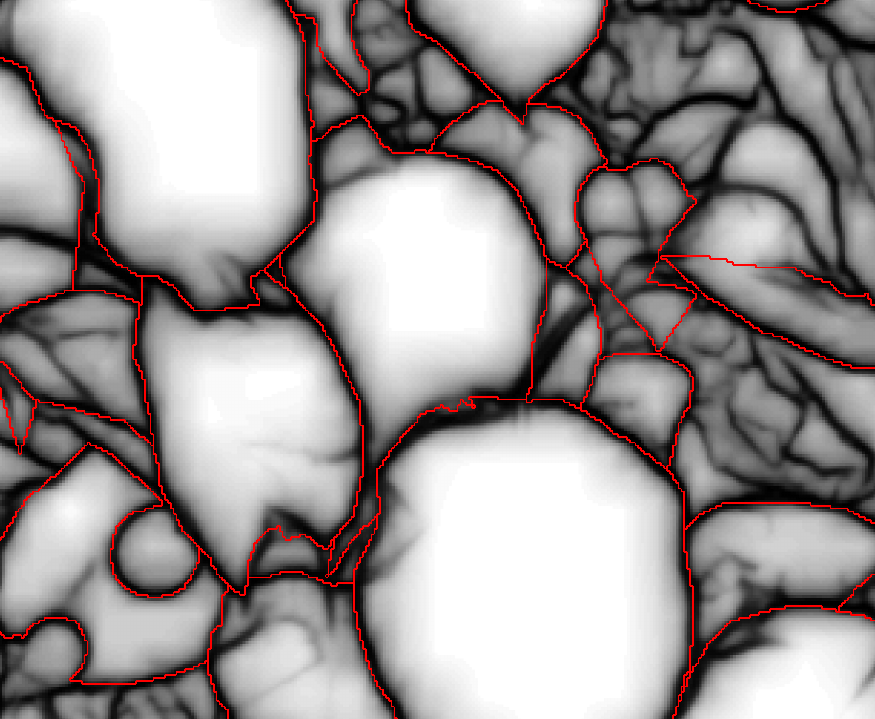}
\subcaption{}
\end{minipage}
\end{center}
\caption{Leaf boundary detection results for the rectangular region in Fig.~\ref{fig:Example1Bounds}.  In each case, darker shows higher probability of boundary, and the red lines are the labeled leaf boundaries.  Ideally the probability of boundary should be zero inside leaves and 1 at the boundaries. (a) Shows a normalized image gradient magnitude.  This is not a good cue when leaves have internal texture. Boundary detection method Pb~\cite{arbelaez:2011} is shown in (b) and HED~\cite{Xie:HED:2017} in (c).  Finally our method, PN, shown in (d), is better able to distinguish true boundaries from internal texture.  Quantitative results are given in Table~\ref{table:results}.    }
\label{fig:Example1Edge}
\end{figure}

\begin{table}[h]
\begin{tabular}{l | c | c | c | c}
\hline
\hline
 Method                             & Boundary AP & Segmentation Dice & Precision & Recall \\
 \hline
 \hline
 PB~\cite{arbelaez:2011}    & 0.866     & 0.599     &	0.511 &	0.724 \\
 \hline
 HED~\cite{Xie:HED:2017}  & 0.852    & 0.728     &	0.668 &	0.800 \\
 \hline
 PN-P1                               & 0.947    & 0.881     &	0.882 &	0.879 \\
 \hline
 PN-P5-S5                          & $\bm{0.960}$    & 0.936     &	0.924 &	$\bm{0.948}$ \\
 \hline
 PN-P5-S1                          & $\bm{0.960}$    & $\bm{0.952}$     &	$\bm{0.957}$ &	$\bm{0.948}$ \\
 \hline
 \hline
 \end{tabular}
 \caption{Performance comparisons on Dense-Leaves test dataset.  PN-P1 indicates training with a single prediction output from the 6-layer PN in Fig.~\ref{fig:NetArch}.  PN-P5-S5 has 5 prediction layers and combines these to create the edge prediction.  PN-P5-S1 has 5 prediction layers but uses only the highest resolution prediction for the performing segmentation.  Boundary AP indicates average precision in predicting leaf boundary pixels.  The Dice, precision and recall figures measure pixel predictions for labeled leaf segments, as output by our algorithm.  For HED, these are obtained by applying our segmentation algorithm to the HED edge estimates.    } 
 \label{table:results}
 \end{table}


\begin{figure*}
\begin{center}
\begin{tabular}{ccc}
\includegraphics[bb=0 0 1000 1000,trim=150 310 150 290,clip,width=0.3\linewidth]{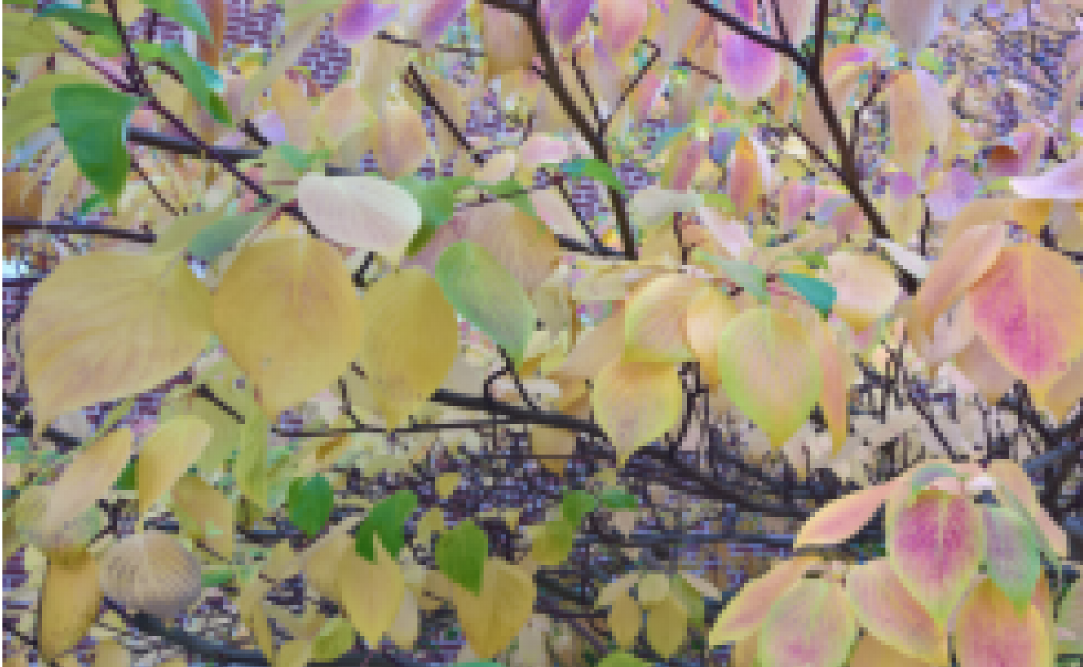} &
\includegraphics[bb=0 0 1000 1000,trim=150 310 150 290,clip,width=0.3\linewidth]{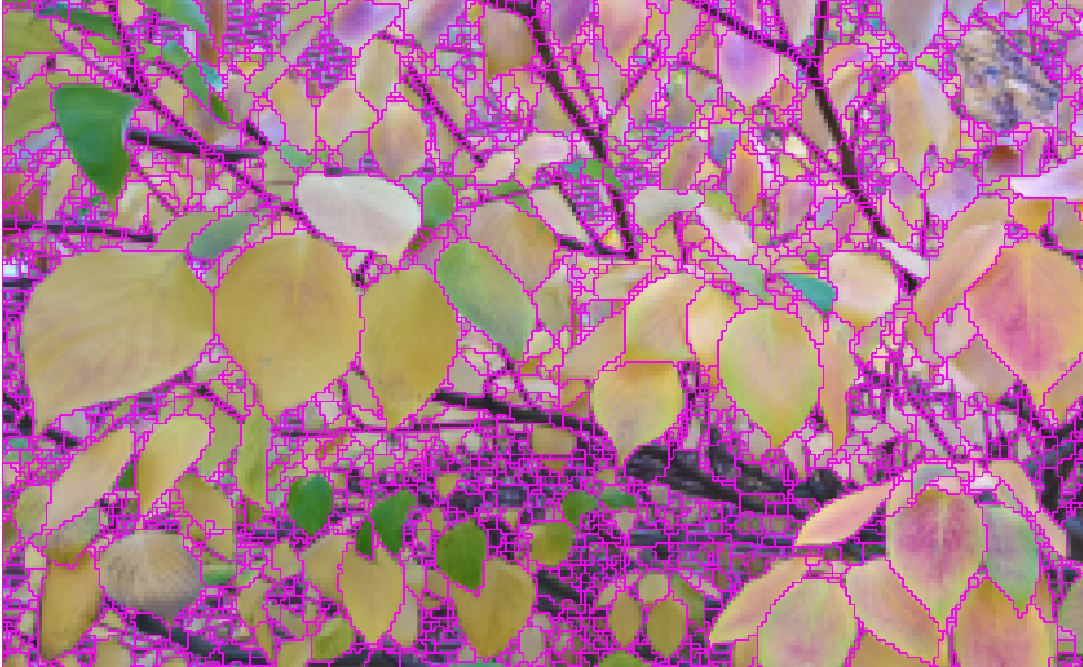} &
\includegraphics[bb=0 0 1000 1000,trim=150 310 150 280,clip,width=0.3\linewidth]{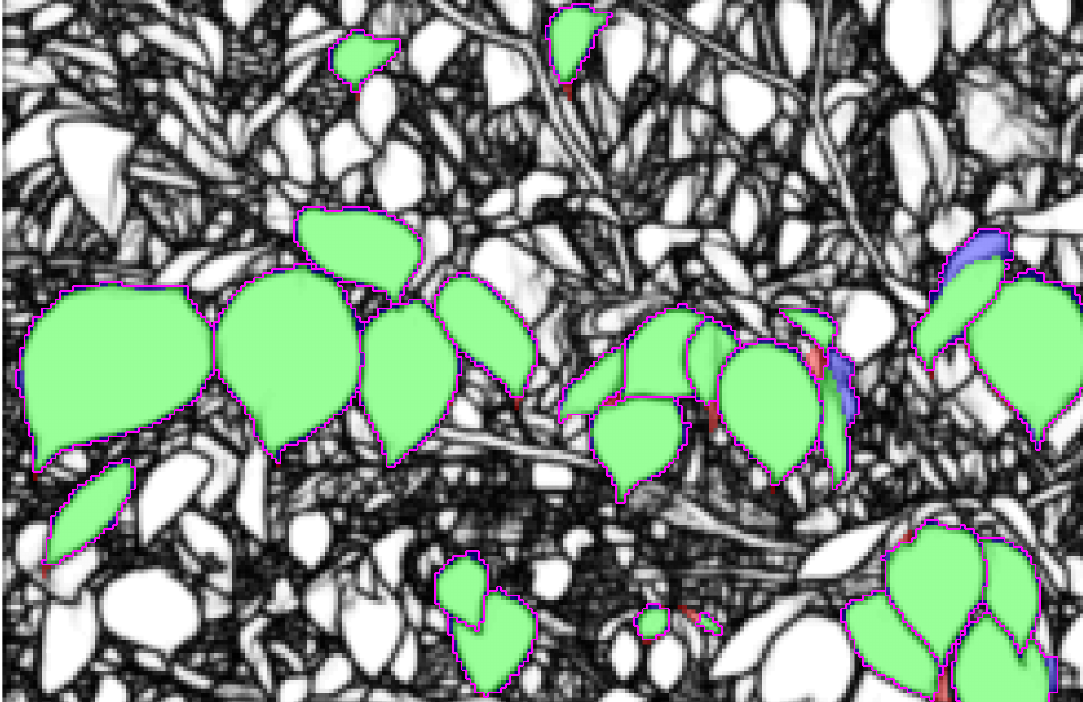} \\
\includegraphics[bb=0 0 1000 1000,trim=150 310 150 290,clip,width=0.3\linewidth]{DenseImage3.pdf} &
\includegraphics[bb=0 0 1000 1000,trim=150 310 150 290,clip,width=0.3\linewidth]{DenseBoundaries3.pdf} &
\includegraphics[bb=0 0 1000 1000,trim=150 310 150 280,clip,width=0.3\linewidth]{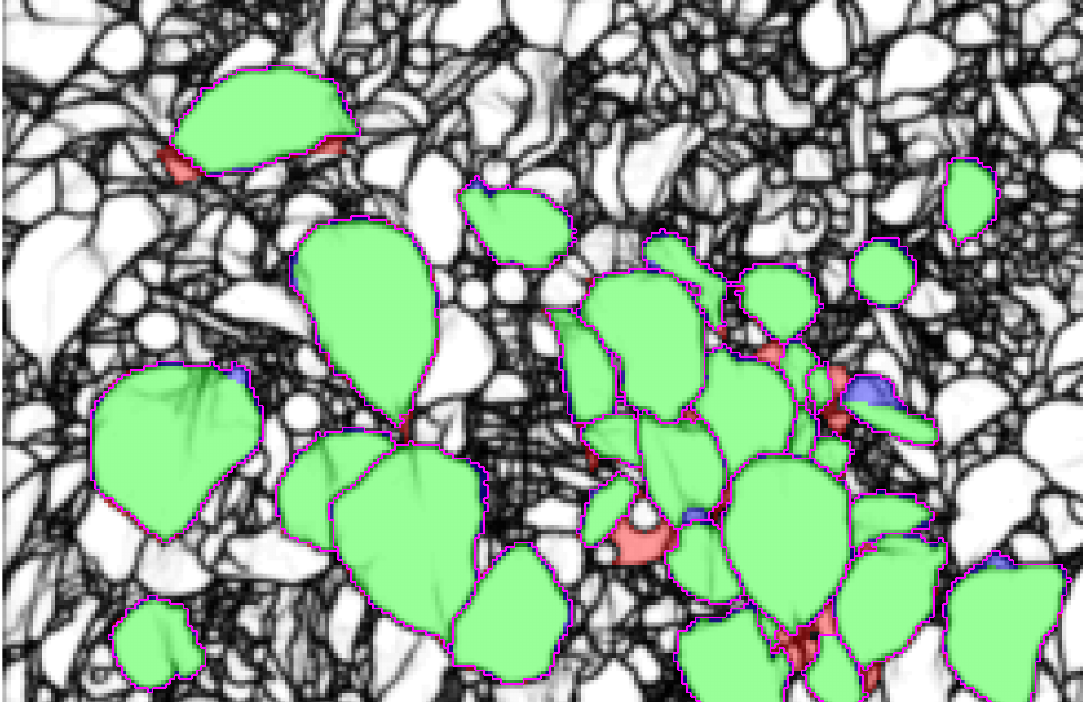} \\
\includegraphics[bb=0 0 1000 1000,trim=150 310 150 290,clip,width=0.3\linewidth]{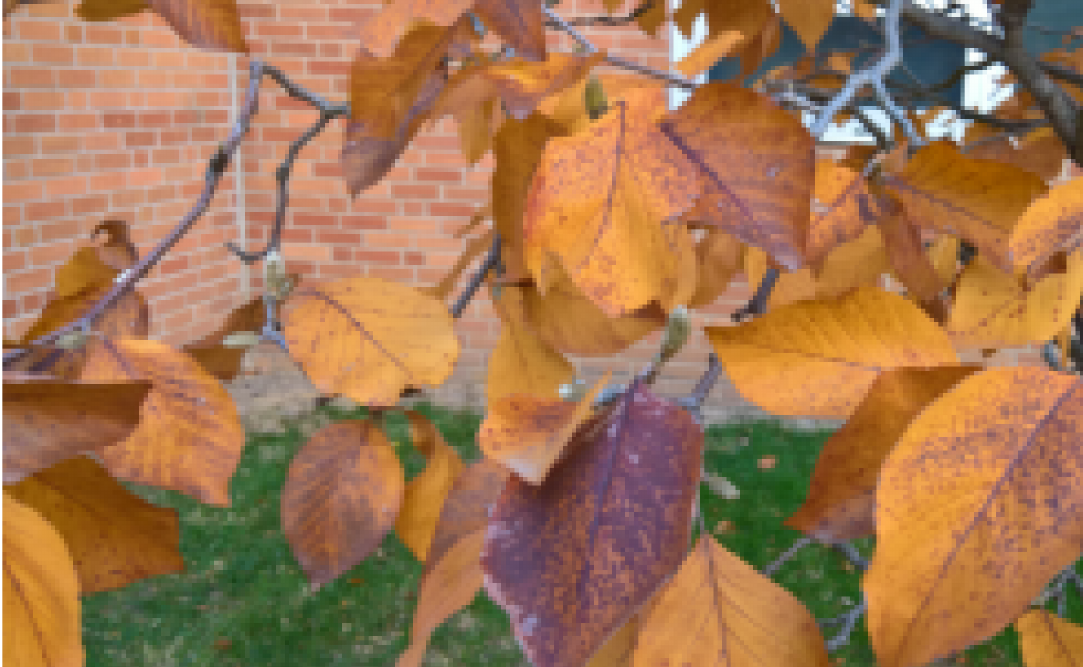} &
\includegraphics[bb=0 0 1000 1000,trim=150 310 150 290,clip,width=0.3\linewidth]{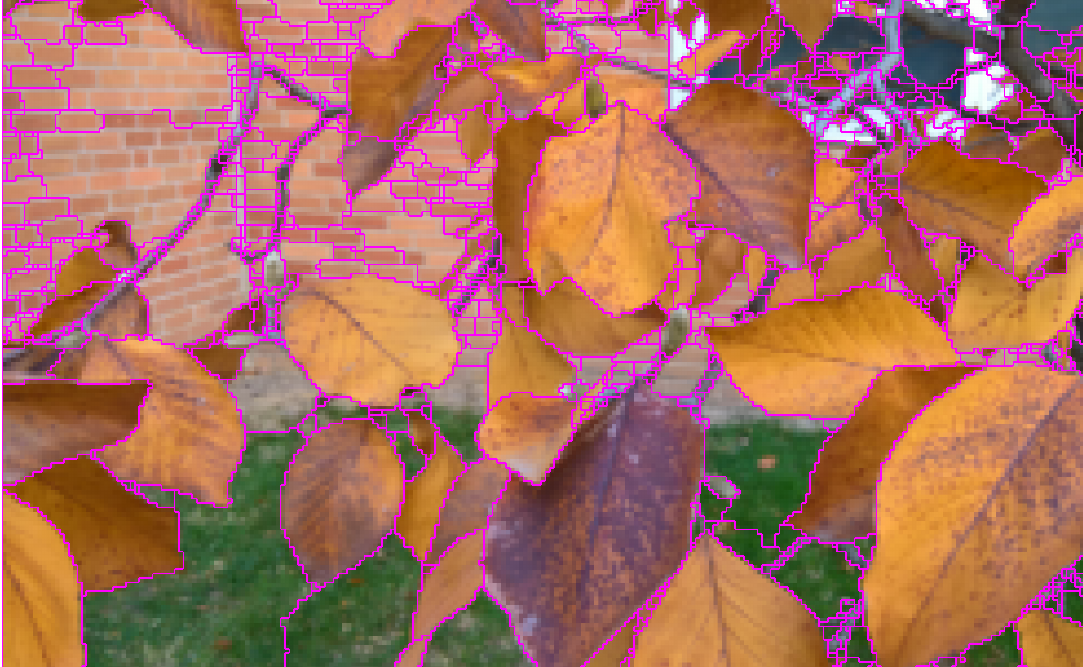} &
\includegraphics[bb=0 0 1000 1000,trim=150 310 150 290,clip,width=0.3\linewidth]{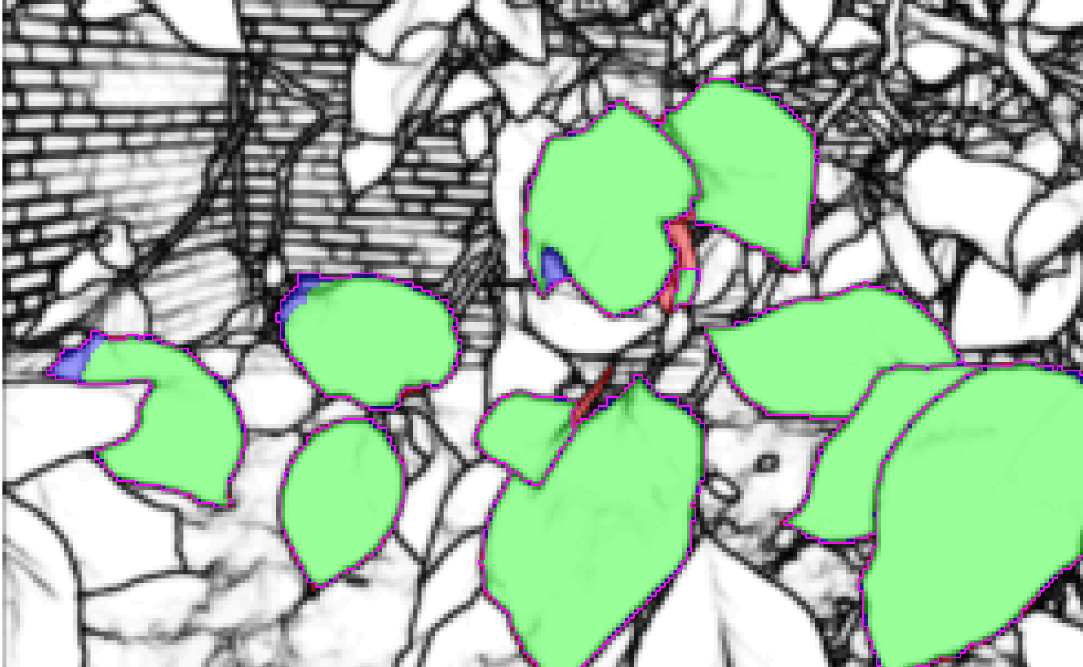} \\
\includegraphics[bb=0 0 1000 1000,trim=150 310 150 290,clip,width=0.3\linewidth]{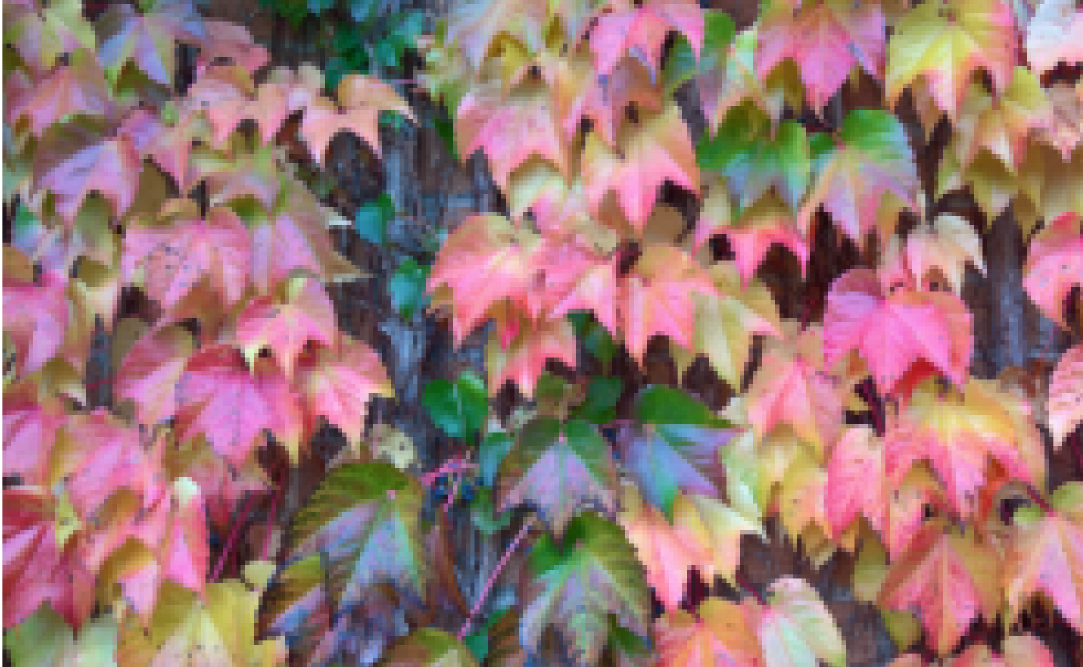} &
\includegraphics[bb=0 0 1000 1000,trim=150 310 150 290,clip,width=0.3\linewidth]{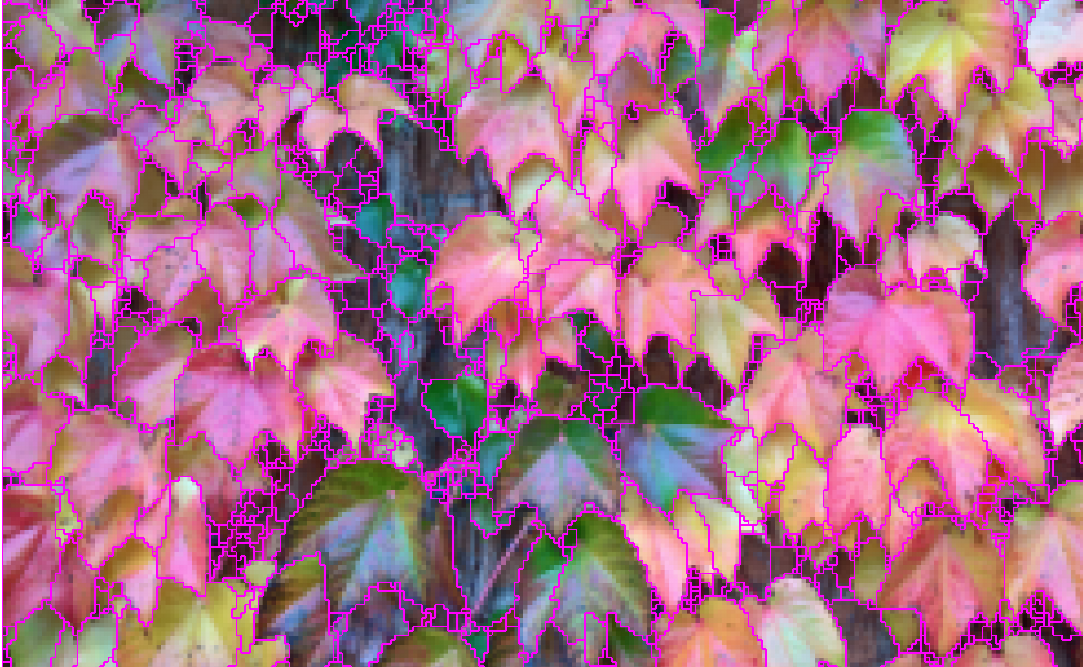} &
\includegraphics[bb=0 0 1000 1000,trim=150 310 150 280,clip,width=0.3\linewidth]{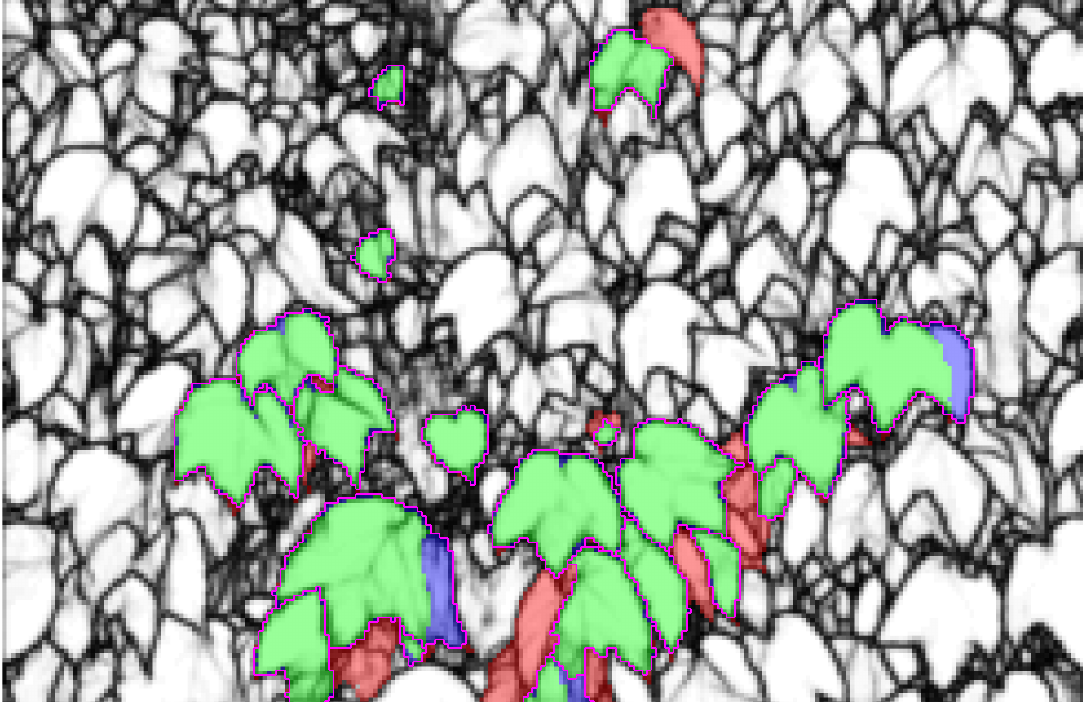} \\
\includegraphics[bb=0 0 1000 1000,trim=150 310 150 290,clip,width=0.3\linewidth]{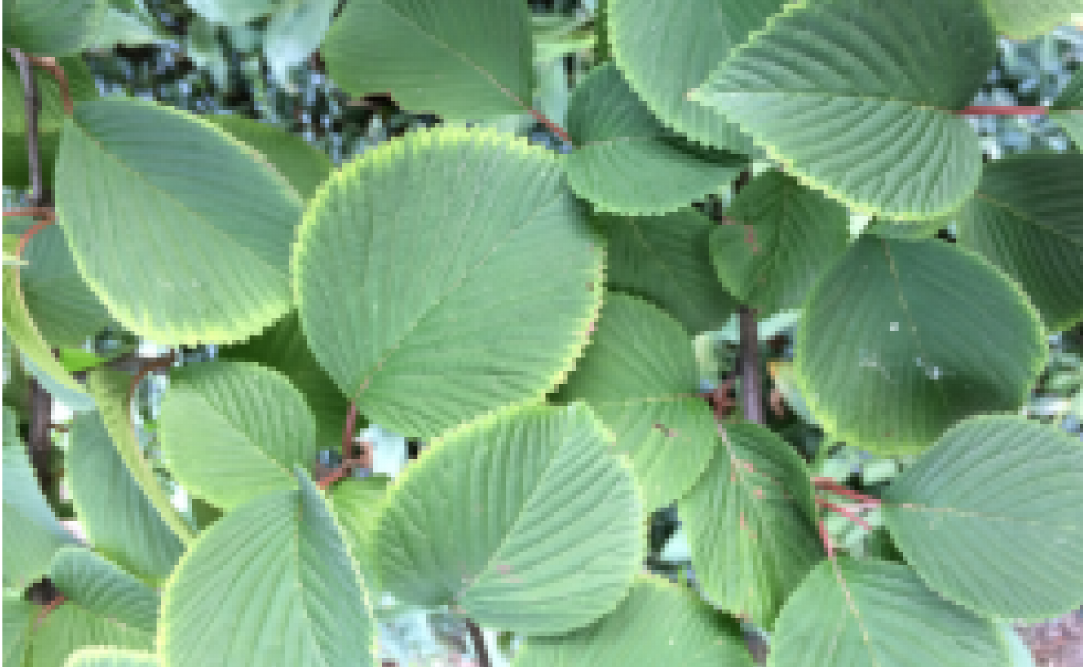} &
\includegraphics[bb=0 0 1000 1000,trim=150 310 150 290,clip,width=0.3\linewidth]{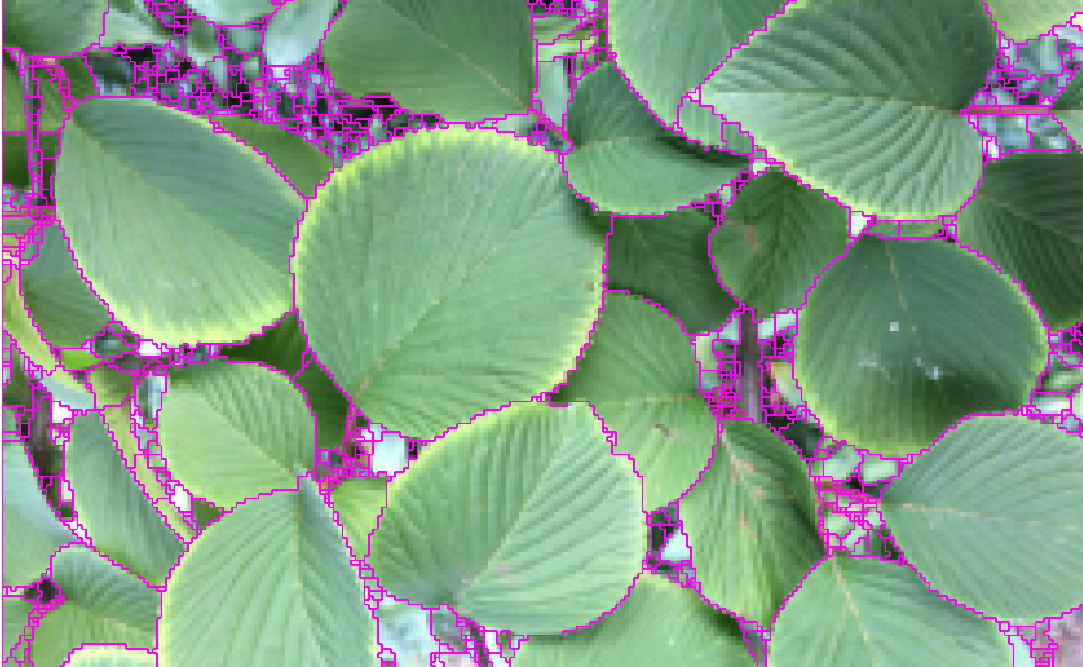} &
\includegraphics[bb=0 0 1000 1000,trim=150 310 150 280,clip,width=0.3\linewidth]{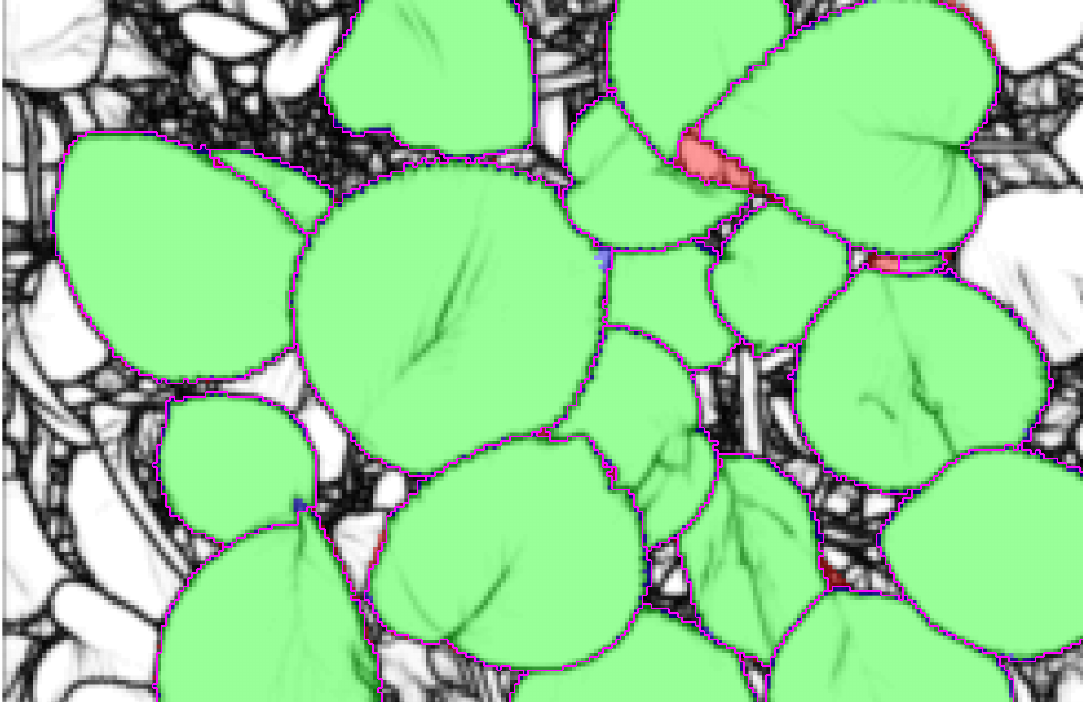} \\
(a) & (b) & (c) \\
\end{tabular}
\caption{(a) The left column shows images from the test set. (b) Center column is the final result of building closed contours from the PN boundaries on the images.  (c) The right column shows boundary detections in black.  In addition the detected segments can be quantified by their overlap with ground truth labeled segments (for the labeled portions of the images).  For each ground truth segment a single best estimated segment is selected.  The overlapping pixels for this segment are true positives (green).  False negatives (blue) are pixels in the ground truth segment pixels not covered by the selected segment, and false positives (red) are pixels of the selected segment that extend outside the corresponding ground truth segment. }
\label{fig:Results}
\end{center}
\end{figure*}

A sample of the boundary detections, final segmentations, and comparisons with ground truths in the test set are shown in Fig.~\ref{fig:Results}.  Most of the internal leaf texture is ignored by the boundary detector, with some exceptions, and most of the boundaries are recovered leading to generally good leaf segmentations.

Our leaf segmentation method can be run on the tree leaf dataset~\cite{GrandBrochier:2015:treeLeaves}.  One modification we made was to reduce the image dimensions by half since those leaves are much higher resolution than the leaves in our dataset.  The only practical difference is that this reduces the accuracy of the boundary.    It is not exactly a fair comparison as our method does not use any leaf vs. background information (namely that the leaf is in the center of the image).    Nevertheless our method achieves dice, precision and recall scores of 0.915, 0.983, 0.856 respectively that outperforms the best method reported there, \cite{cerutti2013}, which has 0.881, 0.927, 0.860 respectively for these scores.   

\section{Conclusion}
\label{sec:conclusion}

This paper proposed PN, a fully convolutional neural network designed and trained to discriminate leaf boundaries.  Using these boundary predictions, a method to generate closed-boundary leaf segmentations is also proposed.  This segmentation task is particularly challenging for leaves with internal texture.  So, to aid in this task and encourage additional work, a labeled dataset of leaves in dense foliage is presented.  Results on this dataset are very promising.  Future work will focus on improving performance for leaves with weak boundary cues.  Part of this will be to significantly expand the dataset to enable training of deeper models.  

\section*{Acknowledgements}
This research was in part supported by a MSU startup grant.  The author expresses his appreciation for the labeling performed by Vivian Morris and Joelle Morris.

{\small

}

\end{document}